\documentclass[11pt,a4paper]{article}
\usepackage[hyperref]{acl2019}
\usepackage{times}
\usepackage{latexsym}

\usepackage{url}

\usepackage{diagbox}
\usepackage{multirow}
\usepackage{bm}
\usepackage{graphicx}
\usepackage{amsmath}
\usepackage{amsfonts}
\usepackage{tabularx}
\usepackage{stfloats}
\usepackage{subcaption}
\usepackage[export]{adjustbox}

\usepackage{xcolor}

\aclfinalcopy 

\title{Few-Shot Dialogue Generation Without Annotated Data: A Transfer Learning Approach}

\author{Igor Shalyminov\textsuperscript{\dag}, Sungjin Lee\textsuperscript{\ddag}, Arash Eshghi\textsuperscript{\dag}, and Oliver Lemon\textsuperscript{\dag} \\
  \textsuperscript{\dag}Heriot-Watt University, UK \\
  \textsuperscript{\ddag}Microsoft Research, US \\
  \texttt{\textsuperscript{\dag}\{is33, a.eshghi, o.lemon\}@hw.ac.uk, \textsuperscript{\ddag}sungjinlee.plus@gmail.com}
  }

\date{}

\begin{document}
\maketitle
\begin{abstract}
Learning with minimal data is one of the key challenges in the development of practical, production-ready goal-oriented dialogue systems. In a real-world enterprise setting where dialogue systems are developed rapidly and are expected to work robustly for an ever-growing variety of domains, products, and scenarios, efficient learning from a limited number of examples becomes indispensable.
  
In this paper, we introduce a technique to achieve state-of-the-art dialogue generation performance in a few-shot setup, without using any annotated data. We do this by leveraging background knowledge from a larger, more highly represented dialogue source~---namely, the MetaLWOz dataset. We evaluate our model on the Stanford Multi-Domain Dialogue Dataset, consisting of human-human goal-oriented dialogues in in-car navigation, appointment scheduling, and weather information domains.
  
We show that our few-shot approach achieves state-of-the art results on that dataset by consistently outperforming the previous best model in terms of BLEU and Entity F1 scores, while being more data-efficient by not requiring any data annotation.

\end{abstract}

\section{Introduction}
\label{sec:intro}
Data-driven dialogue systems are becoming widely adopted in enterprise environments. One of the key properties of a dialogue model in this setting is its {\it data efficiency}, i.e. whether it can attain high accuracy and good generalization properties when only trained from minimal data.

Recent deep learning-based approaches to training dialogue systems \cite{DBLP:conf/sigdial/UltesBCRTWYG18,DBLP:conf/icml/WenMBY17}
put emphasis on collecting large amounts of data in order to account for numerous variations in the user inputs and to cover as many dialogue trajectories as possible. However, in real-world production environments there isn't enough domain-specific data easily available throughout the development process. In addition, it's important to be able to rapidly adjust a system's behavior according to updates in requirements and new product features in the domain. Therefore, data-efficient training is a priority direction in dialogue system research.

In this paper, we build on a technique to train a dialogue model for a new domain in a `zero-shot' setup (in terms of full dialogues in the target domain) only using \emph{annotated} `seed' utterances \cite{DBLP:conf/sigdial/ZhaoE18}.

We present an alternative, `few-shot' approach to data-efficient dialogue system training: we do use complete in-domain dialogues while using approximately the same amount of training data as \newcite{DBLP:conf/sigdial/ZhaoE18}, with respect to utterances. However, in our method, \emph{no annotation is required} ~--- we instead use a latent dialogue act annotation learned in an unsupervised way from a larger (multi-domain) data source, broadly following the model of \newcite{DBLP:conf/acl/EskenaziLZ18}. This approach is potentially more attractive for practical purposes because it is easier to collect unannotated dialogues than collecting utterances across various domains under a consistent annotation scheme. 



\section{Related Work}
\label{sec:related}

There is a substantial amount of work on learning dialogue with minimal data~--- starting with the Dialog State Tracking Challenge 3 \cite{DBLP:conf/slt/HendersonTW14} where the problem was to adjust a pre-trained state tracker to a different domain using a seed dataset.

In dialogue response generation, there has also been work on  bootstrapping a goal-oriented dialogue system from a few examples using a linguistically informed model: \cite{DBLP:conf/emnlp/EshghiSL17} used an incremental semantic parser -- DyLan \cite{Eshghi.etal11,Eshghi15} -- to obtain contextual meaning representations, and based the dialogue state on this \cite{Kalatzis.etal16}. Incremental response generation was learned using Reinforcement Learning, again using the parser to incrementally process the agent's output and thus prune ungrammatical paths for the learner. Compared to a neural model~--- End-to-End Memory Network \cite{DBLP:conf/nips/SukhbaatarSWF15}, this linguistically informed model was superior in a 1-shot setting \cite{DBLP:journals/corr/abs-1709-07840}. At the same time, its main linguistic resource~--- a domain-general dialogue grammar for English~--- makes the model inflexible unless wide coverage is achieved.


Transfer learning for Natural Language Processing is strongly motivated by recent advances in vision. When training a convolutional neural network (CNN) on a small dataset for a specific problem domain, it often helps to learn low-level convolutional features from a greater, more diverse dataset. For numerous applications in vision, ImageNet \cite{imagenet_cvpr09} became the source dataset for pre-training convolutional models. For NLP, the main means for transfer were Word2Vec word embeddings \cite{DBLP:conf/nips/MikolovSCCD13} which have recently been  updated to models capturing contexts as well \cite{DBLP:conf/naacl/PetersNIGCLZ18, DBLP:journals/corr/abs-1810-04805}. While these tools are widely known to improve performance in various tasks, more specialized models could as well be created for specific research areas, e.g. dialogue generation in our case. 

The models above are some of the approaches to one of the central issues of efficient knowledge transfer~--- learning a  unified data representation generalizable across datasets, dubbed `representation learning'. In our approach, we will use one such technique based on variational autoencoding with discrete latent variables \cite{DBLP:conf/acl/EskenaziLZ18}.
In this paper we present an approach to transfer learning which is  more tailored~--- both model-wise and dataset-wise~--- to goal-oriented dialogue in underrepresented domains.



\section{The approach}
\label{sec:approach}

\begin{figure*}
\centering
\begin{subfigure}[b]{.43\textwidth}
  \includegraphics[width=0.95\linewidth,left]{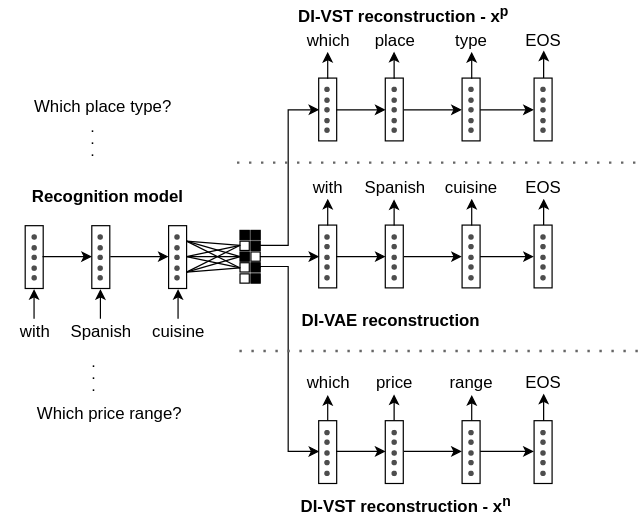}
  \caption{LAED pre-training}
  \label{fig:fsdg_stage1}
\end{subfigure}%
\begin{subfigure}[b]{.57\textwidth}
  \includegraphics[width=0.95\linewidth,right]{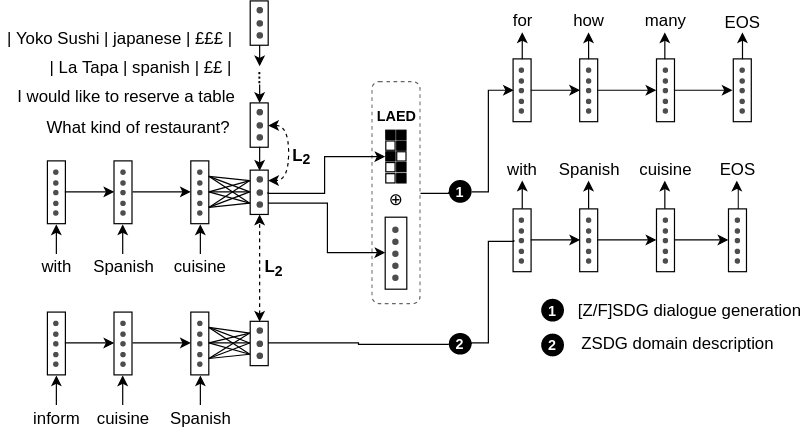}
  \caption{Zero/few-shot dialogue generation}
  \label{fig:fsdg_stage2}
\end{subfigure}
\caption{Model architecture. At the pre-training stage (\ref{fig:fsdg_stage1}), we train the discretized LAED dialogue representation on the Transfer dataset. We then train a zero/few-shot dialogue generation model on SMD with this representation incorporated (\ref{fig:fsdg_stage2}).}
\label{fig:fsdg}
\end{figure*}

\subsection{Zero-shot theoretical framework}
We first describe the original Zero-Shot Dialogue Generation (ZSDG) theoretical framework of \cite{DBLP:conf/sigdial/ZhaoE18} which we base our work on. For ZSDG, there is a set of source dialogue domains and one target domain, with the task of training a dialogue response generation model from all the available source data and a significantly reduced subset of the target data (referred to as \textit{seed} data). The trained system's performance is evaluated exclusively on the target domain.

More specifically, the data in ZSDG is organized as follows. There are unannotated dialogues in the form of $\{\bm{c}, \bm{x}, d\}_{src/tgt}$~--- tuples of dialogue contexts, responses, and domain names respectively for each of the source and target domains. There are also domain descriptions in the form of $\{\bm{x}, \bm{a}, d\}_{src/tgt}$~--- tuples of utterances, slot-value annotations, and domain names respectively for source and target domains.

ZSDG is essentially a hierarchical encoder-decoder model which is trained in a multi-task fashion by 
receiving two types of data: (1) dialogue batches drawn from all the available source-domain data, and (2) seed data batches, a limited number of which are drawn from domain description data for all of the source and target domains.

ZSDG model optimizes for 2 objectives. With dialogue batches, the model maximizes the probability of generating a response given the context:
\begin{equation}
\label{eq:zsdg_l_dialog}
\begin{split}
  \mathcal{L}_{dialog} = & -\log p_{\mathcal{F}^d}(\mathbf{x} \mid \mathcal{F}^e(\mathbf{c}, d)) \\
  &+ \lambda \mathcal{D}(\mathcal{R}(\mathbf{x}, d) \rVert \mathcal{F}^e(\mathbf{c}, d))
\end{split}
\end{equation}
where $\mathcal{F}^e$ and $\mathcal{F}^d$ are respectively the encoding and decoding components of a hierarchical generative model; $\mathcal{R}$ is the shared recurrent utterance encoder (the \textit{recognition model}); and $\mathcal{D}$ is a distance function ($L2$ norm).

In turn, with domain description batches, the model maximizes the probability of generating the utterance given its slot-value annotation, both represented as sequences of tokens:
\begin{equation}
\label{eq:zsdg_l_dd}
\begin{split}
  \mathcal{L}_{dd} = & -\log p_{\mathcal{F}^d}(\mathbf{x} \mid \mathcal{R}(\mathbf{a}, d)) \\
  & + \lambda \mathcal{D}(\mathcal{R}(\mathbf{x}, d) \rVert \mathcal{R}(\mathbf{a}, d))
\end{split}
\end{equation}

In this multi-task setup, the latent space of $\mathcal{R}$ is shared between both utterances and domain descriptions across all the domains. Moreover, the distance-based loss terms make sure that (a) utterances with similar annotations are closer together in the latent space (Eq. \ref{eq:zsdg_l_dd}), and (b) utterances are closer to their dialogue contexts (Eq. \ref{eq:zsdg_l_dialog}) so that their encodings capture some of the contexts' meaning. These properties of the model make it possible to achieve better cross-domain generalization.

\subsection{Unsupervised representation learning}
As was the case with ZSDG, robust representation learning helps achieve better generalization across domains. The most widely-adopted way to train better representations has been to leverage a greater data source.
In this work, we consider unsupervised, variational autoencoder-based (VAE) representation learning on a large dataset of unannotated dialogues. The specific approach we refer to is the Latent Action Encoder-Decoder (LAED) model of \cite{DBLP:conf/acl/EskenaziLZ18}. LAED is a variant of VAE with two modifications: (1) an optimization objective augmented with mutual information between the input and the latent variable for better and more stable learning performance,  and (2) discretized latent variable for the interpretability of the resulting latent actions. Just as in ZSDG, LAED is a hierarchical encoder-decoder model with the key component being a discrete-information (DI) utterance-level VAE. Two versions of this model are introduced, with respective optimization objectives:

\begin{equation}
\label{eq:di_vae}
\begin{split}
  \mathcal{L}_{DI\text{-}VAE} = & \mathbb{E}_{q_{\mathcal{R}}(\bm{z} \mid \bm{x}) p(\bm{x})} [ \log p_\mathcal{G}(\bm{x} \mid \bm{z})] \\
  & - KL(q(\bm{z}) \rVert p(\bm{z}))
\end{split}
\end{equation}

\begin{equation}
\label{eq:di_vst}
\begin{split}
  \mathcal{L}_{DI\text{-}VST} & = \mathbb{E}_{q_{\mathcal{R}}(\bm{z}\mid\bm{x}) p(\bm{x})} [ \log p^n_\mathcal{G}(\bm{x_n} \mid \bm{z}) p^p_\mathcal{G}(\bm{x_p} \mid \bm{z})] \\
  & - KL(q(\bm{z}) \rVert p(\bm{z}))
\end{split}
\end{equation}

where $\mathcal{R}$ and $\mathcal{G}$ are recognition and generation components respectively, $\bm{x}$ is the model's input, $\bm{z}$ is the latent variable, and $p(\bm{z})$ and $q(\bm{z})$ are respectively prior and posterior distributions of $\bm{z}$.

DI-VAE works in a standard VAE fashion reconstructing the input $\bm{x}$ itself, while DI-VST follows the idea of Variational Skip-Thought reconstructing the input's previous and next contexts: $\{\bm{x_n}, \bm{x_p}\}$. As reported by the authors, the two models capture different aspects of utterances, i.e. DI-VAE reconstructs specific words within an utterance, whereas DI-VST captures the overall intent better~--- see the visualization in Figure \ref{fig:fsdg_stage1}.

\subsection{Proposed models\footnote{Code is available at \url{https://bit.ly/fsdg_sigdial2019}}}

\begin{table*}[t!]
  \footnotesize
  \centering
    \begin{tabular}{|l||l|l||l|l||l|l|}
      \hline
      \multirow{2}{*}{\backslashbox{\textbf{Model}}{\textbf{Domain}}}&\multicolumn{2}{c||}{\textbf{Navigation}}&\multicolumn{2}{c||}{\textbf{Weather}}&\multicolumn{2}{c|}{\textbf{Schedule}}\\
      &\multicolumn{1}{c|}{BLEU, \%}&\multicolumn{1}{c||}{Entity F1, \%}&\multicolumn{1}{c|}{BLEU, \%}&\multicolumn{1}{c||}{Entity F1, \%}&\multicolumn{1}{c|}{BLEU, \%}&\multicolumn{1}{c|}{Entity F1, \%}\\\hline\hline
      ZSDG&5.9&14.0&8.1&31&7.9&36.9\\
      NLU\_ZSDG&$6.1\pm2.2$&$12.7\pm3.3$&$5.0\pm1.6$&$16.8\pm6.7$&$6.0\pm1.7$&$26.5\pm5.4$\\
      NLU\_ZSDG+LAED&$7.9\pm1$&$12.3\pm2.9$&$8.7\pm0.6$&$21.5\pm6.2$&$8.3\pm1$&$20.7\pm4.8$\\
      \hline
      FSDG@1\%&$6.0\pm1.8$&$9.8\pm4.8$&$6.9\pm1.1$&$22.2\pm 10.7$&$5.5\pm0.8$&$25.6\pm8.2$\\
      FSDG@3\%&$7.9\pm0.7$&$11.8\pm4.4$&$9.6\pm1.8$&$39.8\pm 7$&$8.2\pm1.1$&$34.8\pm4.4$\\
      FSDG@5\%&$8.3\pm1.3$&$15.3\pm6.3$&$11.5\pm 1.6$&$38.0\pm10.5$&$9.7\pm1.4$&$37.6\pm8.0$\\
      FSDG@10\%&$9.8\pm0.8$&$19.2\pm3.2$&$12.9\pm2.4$&$40.4\pm11.0$&$12.0\pm1.0$&$38.2\pm4.2$\\
      \hline
      FSDG+VAE@1\%&$3.6\pm2.6$&$9.3\pm4.1$&$6.8\pm1.3$&$23.2\pm10.1$&$4.6\pm1.6$&$28.9\pm7.3$\\
      FSDG+VAE@3\%&$6.9\pm1.9$&$15.6\pm5.8$&$9.5\pm2.6$&$32.2\pm11.8$&$6.6\pm1.7$&$34.8\pm7.7$\\
      FSDG+VAE@5\%&$7.8\pm1.9$&$12.7\pm4.2$&$10.1\pm2.1$&$40.3\pm10.4$&$8.2\pm1.7$	&$34.2\pm8.7$\\
      FSDG+VAE@10\%&$9.0\pm2.0$&$18.0\pm5.8$&$12.9\pm2.2$&$40.1\pm7.6$&$11.6\pm1.5$&	$39.9\pm6.9$\\\hline
      FSDG+LAED@1\%&$7.1\pm0.8^\mathbf{\star}$&$10.1\pm4.5$&$10.6\pm2.1^\mathbf{\star}$&$31.4\pm8.1^\mathbf{\star}$&$7.4\pm1.2$&$29.1\pm6.6$\\
      FSDG+LAED@3\%&$9.2\pm0.8$&$14.5\pm4.8^\mathbf{\star}$&$13.1\pm1.7$&$40.8\pm6.1$&$9.2\pm1.2^\mathbf{\star}$&$32.7\pm6.1$\\
      \textbf{FSDG+LAED@5\%}&$\bm{10.3\pm1.2}$&$\bm{15.6\pm4.5}$&$\bm{14.5\pm2.2}$&$\bm{40.9\pm8.6}$&$\bm{11.8\pm1.9}$&$\bm{37.6\pm6.1^*}$\\
      FSDG+LAED@10\%&$12.3\pm0.9$&$17.3\pm4.5$&$17.6\pm1.9$&$47.5\pm6.0$&$15.2\pm1.6$	&$38.7\pm8.4$\\\hline
    \end{tabular}
    \caption{Evaluation results. Marked with asterisks are individual results higher than the ZSDG baseline which are achieved with the minimum amount of training data, and in bold is the model consistently outperforming ZSDG in all domains and metrics with minimum data.}
    \label{tab:results}
\end{table*}

In our approach, we simplify the ZSDG setup by not using any explicit domain descriptions, therefore we only work with `dialogue' batches. 
We also make use of Knowledge Base information without loss of generality (see Section \ref{sec:setup})~--- thus we work with data of the form $\{\bm{c}, \bm{x}, \bm{k}, d\}$ where $\bm{k}$ is the KB information. We refer to this model as Few-Shot Dialogue Generation, or \textbf{\textit{FSDG}}.

For learning a reusable dialogue representation, we use an external multi-domain dialogue dataset, the Transfer dataset (see Section \ref{sec:data}).

We perform a preliminary training stage on it where we train 2 LAED models, both DI-VAE and DI-VST. Then, at the main training stage, we use the hierarchical encoders of both models and incorporate them with FSDG's decoder. Thus, we have the following encoding function (which is then plugged in to the Eq. \ref{eq:zsdg_l_dialog}):

\begin{equation}
\label{eq:laed_fsdg_F_e}
\begin{split}
  \mathcal{F}^e(\mathbf{c}, \mathbf{k}, d) & = \mathcal{F}^e_{DI\text{-}VAE}(\mathbf{c}, \mathbf{k}, d) \\
  & \oplus \mathcal{F}^e_{DI\text{-}VST}(\mathbf{c}, \mathbf{k}, d) \\
  & \oplus \mathcal{F}^e_{FSDG}(\mathbf{c}, \mathbf{k}, d)
\end{split}
\end{equation}

where $\oplus$ is the concatenation operator.
We refer to this model as \textbf{\textit{FSDG+LAED}}.

We compare this LAED-augmented model to a similar one, with latent representation trained on the same data but using a regular VAE objective and thus providing regular continuous embeddings (we refer to it as \textbf{\textit{FSDG+VAE}}).

\begin{equation}
\label{eq:vae_fsdg}
\begin{split}
  \mathcal{L}_{VAE} & = \mathbb{E}_{q_{\mathcal{R}}(\bm{z}\mid\bm{x})} [ \log p_\mathcal{G}(\bm{x} \mid \bm{z})] \\
  & - KL(q_{\mathcal{R}}(\bm{z}) \rVert p(\bm{z}))
\end{split}
\end{equation}

Finally, in order to explore the original ZSDG setup as much as possible, we also consider its version with automatic Natural Language Understanding (NLU) markup instead of human annotations as domain descriptions. Our NLU annotations include Named Entity Recognizer \cite{DBLP:conf/acl/FinkelGM05}, a date/time extraction library \cite{DBLP:conf/lrec/ChangM12}, and a Wikidata entity linker \cite{Pappu:WSDM2017}.
We have models with (\textbf{\textit{NLU\_ZSDG+LAED}}) and without LAED representation (\textbf{\textit{NLU\_ZSDG}}). Our entire setup is shown in Figure \ref{fig:fsdg}.

\section{Datasets}
\label{sec:data}

We use the Stanford Multi-Domain (SMD) human-human goal-oriented dialogue dataset \cite{DBLP:conf/sigdial/EricKCM17} in 3 domains: appointment scheduling, city navigation, and weather information. Each dialogue comes with knowledge base snippet from the underlying domain-specific API.

For LAED training, we use MetaLWOz \cite{multi-domain-task-completion-dialog-challenge}, a human-human goal-oriented dialogue corpus specifically designed for various meta-learning and pre-training purposes. It contains conversations in 51 domains with several tasks in each of those. The dialogues are collected using the Wizard-of-Oz method where human participants were given a problem domain and a specific task. No domain-specific APIs or knowledge bases were available for the participants, and in the actual dialogues they were free to use fictional names and entities in a consistent way. The dataset totals more than $40,000$ dialogues, with the average length of $11.9$ turns.

\section{Experimental setup and evaluation}
\label{sec:setup}
Our few-shot setup is as follows. Given the target domain, we first train LAED models (a dialogue-level DI-VST and an utterance-level DI-VAE, both of the size $10\times 5$) on the MetaLWOz dataset~--- here we exclude from training every domain that might overlap with the target one.

Next, using the LAED encoders, we train a Few-Shot Dialogue Generation model on all the SMD source domains. We use a random sample (1\% to 10\%) of the target domain utterances together with their contexts as seed data.

We incorporate KB information into our model by simply serializing the records and prepending them to the dialogue context, ending up with a setup similar to CopyNet in \cite{DBLP:conf/sigdial/EricKCM17}.

For the NLU\_ZSDG setup, we use 1000 random seed utterances from each source domain and 200 utterances from the target domain\footnote{The numbers are selected so that the domain description task is kept secondary.}.

For evaluation, we follow the approach of \cite{DBLP:conf/sigdial/ZhaoE18} and report BLEU and Entity F1 scores~--- means/variances over 10 runs.

\section{Results and discussion}
\label{ref:results}

Our results are shown in Table \ref{tab:results}. Our objective here is maximum accuracy with minimum training data required, and it can be seen that few-shot models with LAED representation are the best performing models for this objective. While the improvements can already be seen with simple FSDG, the use of LAED representation helps to significantly reduce the amount of in-domain training data needed: in most cases, the state-of-the-art results are attained with as little as 3\% of in-domain data. At 5\%, we see that FSDG+LAED consistently improves upon all other models in every domain, either by increasing the mean accuracy or by decreasing the variation. In SMD, with its average dialogue length of 5.25 turns (see Table \ref{tab:smd}), 5\% of training dialogues amounts to  approximately 200 in-domain training utterances. In contrast, the ZSDG setup used approximately 150 {\it annotated} training utterances for each of the 3 domains, totalling about \textit{450 annotated} utterances. Although in our few-shot approach we use full in-domain dialogues, we end up having a comparable amount of target-domain training data, with the crucial difference that none of those has to be annotated for our approach. Therefore, the method we introduced attains state-of-the-art in both accuracy and data-efficiency.

The results of the  ZSDG\_NLU setup demonstrate that single utterance annotations, if not domain-specific and produced by human experts, don't provide as much signal as raw dialogues.

The comparison of the setups with different latent representations also gives us some insight: while the VAE-powered FSDG model improves on the baseline in multiple cases, it lacks generalization potential compared to LAED. The reason for that might be inherently more stable training of LAED due to its modified objective function which in turn results in a more informative, generalizable representation.

Finally, we discuss the evaluation metrics. Since we base this paper on the work of \cite{DBLP:conf/sigdial/ZhaoE18}, we have had to fully conform to the metrics they used to enable direct comparison. However, BLEU as the primary evaluation metric, does not necessarily reflect NLG quality in dialogue settings~--- see examples in Table \ref{tab:examples} of the Appendix (see also \newcite{Novikova.etal17}). This is a general issue in dialogue model evaluation since the variability of possible responses equivalent in meaning is very high in dialogue. In future work, instead of using BLEU, we will  put more emphasis on the meaning of utterances, for example by using external dialogue act tagging resources, using quality metrics of language generation -- e.g. perplexity -- as well as more task-oriented metrics like Entity F1. We expect these to make for more meaningful evaluation criteria.

\section{Conclusion and future work}
\label{ref:future}

In this paper, we have introduced a technique to achieve state-of-the-art dialogue generation performance in a few-shot setup, without using any annotated data. By leveraging larger, more highly represented dialogue sources and learning robust latent dialogue representations from them, we obtained a model with superior generalization to an underrepresented domain. Specifically, we showed that our few-shot approach achieves state-of-the art results on the Stanford Multi-Domain dataset while being more data-efficient than the previous best model, by not requiring any data annotation.

Although being state-of-the-art, the accuracy scores themselves still suggest that our technique is not ready for immediate adoption for real-world production purposes, and the task of few-shot generalization to a new dialogue domain remains an area of active research. We expect that such initiatives will be fostered by the release of large dialogue corpora such as MetaLWOz.

In our own future work, we will try and find ways to improve the unsupervised representation in order to increase the transfer potential. Adversarial learning can also be beneficial in the setting of limited data. And apart from improving the model itself, it is necessary to consider an alternative criterion to BLEU-score for adequate evaluation of response generation.


\bibliography{acl2019,all}
\bibliographystyle{acl_natbib}

\newpage
\clearpage

\begin{table*}[ht!]
\small
\center
\begin{tabular}{lllll}
\hline\hline
\textbf{Domain}&\multicolumn{2}{c}{\textbf{Context}}&\textbf{Gold response}&\textbf{Predicted response}\\\hline
schedule&\texttt{<usr>}&Remind me to take my pills&Ok setting your medicine&Okay, setting \textit{a reminder to take}\\
&\texttt{<sys>}&What time do you need&appointment for 7pm&\textit{your pills at 7 pm.}\\
&&to take your pills?&&\\
&\texttt{<usr>}&I need to take my pills at 7 pm.&&\\
\hline
navigate&\texttt{<usr>}&Find the address to a hospital&Have a good day&\textit{No problem.}\\
&\texttt{<sys>}&Stanford Express Care is&&\\
&&at 214 El Camino Real.&&\\
&\texttt{<usr>}&Thank you.&&\\
\hline
navigate&\texttt{<usr>}&What is the weather forecast&For what city would you&For what city would you like\\
&&for the weekend?&like to know that?&\textit{the weekend forecast for?}\\
\hline\hline
\end{tabular}
\caption{Selected FSDG+LAED model's responses}
\label{tab:examples}
\end{table*}

\begin{table*}[ht!]
\tiny
\centering
\begin{tabular}{|ll||ll||ll|}
\hline
\textbf{Domain}&\textbf{\#Dialogues}&\textbf{Domain}&\textbf{\#Dialogues}&\textbf{Domain}&\textbf{\#Dialogues}\\\hline\hline
UPDATE\_CALENDAR&1991&GUINESS\_CHECK&1886&ALARM\_SET&1681\\
SCAM\_LOOKUP&1658&PLAY\_TIMES&1601&GAME\_RULES&1590\\
CONTACT\_MANAGER&1483&LIBRARY\_REQUEST&1339&INSURANCE&1299\\
HOME\_BOT&1210&HOW\_TO\_BASIC&1086&CITY\_INFO&965\\
TIME\_ZONE&951&TOURISM&935&SHOPPING&903\\
BUS\_SCHEDULE\_BOT&898&CHECK\_STATUS&784&WHAT\_IS\_IT&776\\
STORE\_DETAILS&737&APPOINTMENT\_REMINDER&668&PRESENT\_IDEAS&664\\
GEOGRAPHY&653&SKI\_BOT&607&MOVIE\_LISTINGS&607\\
UPDATE\_CONTACT&581&ORDER\_PIZZA&577&EDIT\_PLAYLIST&574\\
SPORTS\_INFO&561&BOOKING\_FLIGHT&555&WEATHER\_CHECK&551\\
EVENT\_RESERVE&539&RESTAURANT\_PICKER&535&LOOK\_UP\_INFO&533\\
AUTO\_SORT&514&QUOTE\_OF\_THE\_DAY\_BOT&513&WEDDING\_PLANNER&510\\
MAKE\_RESTAURANT\_RESERVATIONS&510&AGREEMENT\_BOT&507&NAME\_SUGGESTER&499\\
APARTMENT\_FINDER&499&HOTEL\_RESERVE&497&PHONE\_PLAN\_BOT&496\\
DECIDER\_BOT&487&VACATION\_IDEAS&486&PHONE\_SETTINGS&473\\
POLICY\_BOT&447&PROMPT\_GENERATOR&446&MUSIC\_SUGGESTER&445\\
PET\_ADVICE&426&BANK\_BOT&367&CATALOGUE\_BOT&288\\\hline
\end{tabular}
\caption{MetaLWOz domains}
\label{tab:maluuba_domains}
\end{table*}

\appendix

\section{Appendices}
\label{sec:appendix}

\subsection{Training details}
\label{sec:apx_training}
We train our models with the Adam optimizer \cite{DBLP:journals/corr/KingmaB14} with learning rate $0.001$. Our hierarchical models' utterance encoder is an LSTM cell \cite{DBLP:journals/neco/HochreiterS97} of   size $256$, and the dialogue-level encoder is a GRU \cite{DBLP:conf/emnlp/ChoMGBBSB14} of   size $512$.

\begin{table}[h!]
\footnotesize
\center
\begin{tabular}{|l||l|l|l|}
\hline
\backslashbox{\textbf{Statistic}}{\textbf{Domain}}&Navigation&Weather&Schedule\\\hline\hline
Dialogues&800&797&8285\\\hline
Utterances &5248&4314&3170\\\hline
Mean length&6.56&5.41&3.83\\\hline
\end{tabular}
\caption{SMD trainset statistics}
\label{tab:smd}
\end{table}

\subsection{MetaLWOz dialogue example}
\textbf{Domain}: Ski \\
\textbf{User Task}: You want to know if there are good ski hills an hour's drive from your current location. \\
\textbf{Bot Task}: Tell the user that there are no ski hills in their immediate location. \\
\textcolor{red}{Bot} Hello how may I help you? \\
\textcolor{blue}{User} Is there any good ski hills an hour's drive from my current location? \\
\textcolor{red}{Bot} I'm sorry to inform you that there are no ski hills in your immediate location \\
\textcolor{blue}{User} Can you help me find the nearest? \\
\textcolor{red}{Bot} Absolutely! It looks like you're about 3 hours away from Bear Mountain. That seems to be the closest.\\
\textcolor{blue}{User} Hmm.. sounds good \\
\textcolor{red}{Bot} Alright! I can help you get your lift tickets now! When will you be going? \\
\textcolor{blue}{User} Awesome! please get me a ticket for 10pax \\
\textcolor{red}{Bot} You've got it. Anything else I can help you with? \\
\textcolor{blue}{User} None. Thanks again! \\
\textcolor{red}{Bot} No problem! \\

\end{document}